\documentclass{article}
\usepackage{microtype}
\usepackage{graphicx}
\usepackage{booktabs} 

\usepackage{hyperref}

\usepackage{amsmath}
\usepackage{amssymb}
\usepackage{mathtools}
\usepackage{amsthm}

\usepackage{hyperref}
\usepackage{url}
\usepackage{graphicx}
\usepackage{times}
\usepackage{microtype}
\usepackage{graphicx}
\usepackage{booktabs} 
\usepackage{multicol}
\usepackage{amssymb}
\usepackage{mathtools}
\usepackage{soul}
\usepackage{booktabs}
\usepackage{siunitx}
\usepackage{multirow}
\usepackage{xcolor}
\usepackage{enumitem}
\usepackage{lipsum}
\usepackage{breqn}
\usepackage{booktabs}
\usepackage{siunitx}
\usepackage{multirow}
\usepackage{graphicx}
\usepackage[font=small]{caption}
\usepackage{lipsum}
\usepackage{enumitem}
\usepackage{multicol}
\usepackage{amsthm}
\usepackage{amsmath,amssymb,mathtools}
\usepackage{booktabs}
\usepackage{siunitx}
\usepackage{multirow}
\usepackage{xcolor}
\usepackage{soul}
\usepackage{float}
\usepackage{pifont}

\usepackage[final]{corl_2021} 
\usepackage{setspace}

\title{Real Robot Challenge 2021: Cartesian Position Control with Triangle Grasp and Trajectory Interpolation}

\author{
  Rishabh Madan\\
  Cornell University\\ 
  \texttt{rm773@cornell.edu} \\
   \And
   Harshit Sikchi \\
   The University of Texas at Austin \\
   \texttt{hsikchi@utexas.edu} \\
   \AND
   Ethan K. Gordon \\
   University of Washington \\
   \texttt{ekgordon@cs.washington.edu} \\
   \And
   Tapomayukh Bhattacharjee \\
   Cornell University \\
   \texttt{tapomayukh@cornell.edu} \\
}

\begin{document}
\maketitle


\begin{abstract}
    We present our runner-up approach for the Real Robot Challenge 2021. We build upon our previous approach used in Real Robot Challenge 2020~\cite{funk2021benchmarking}. To solve the sequential goal-reaching task introduced in RRC 2021, we focus on two aspects to achieving near-optimal trajectory: Grasp stability and Controller performance. In the RRC 2021 simulated challenge, our method utilizes a hand-designed Three-Jaw Chuck Grasp combined with Trajectory Interpolation for better stability during the motion for fast goal reaching. In Stage 1, we observe reverting to Triangle Grasp, which was used in the 2020 Edition, provides a more stable grasp when combined with Trajectory Interpolation. The video demonstration for our approach is available at \href{https://youtu.be/dlOueoaRWrM}{https://youtu.be/dlOueoaRWrM}. The code is publicly available at \href{https://github.com/madan96/benchmark-rrc}{https://github.com/madan96/benchmark-rrc}.
\end{abstract}



\section{Introduction}

Dexterous manipulation, although an everyday task for humans, is a challenging task in the robotics community. Some of the prior successes~\cite{nagabandi2020deep,andrychowicz2020learning,inhandturning, gu2017deep} come with prohibitive engineering costs. Real Robot Challenge (RRC) presents a testbed for roboticists worldwide to remotely experiment with their method on robust real-world robotic systems and thus promotes democratization of robotics, particularly robot manipulation. The environment in Stage 1 of RRC 2021\footnote{https://real-robot-challenge.com/protocol} consists of a three-finger manipulator tasked with grasping a cube and moving it in a sequence of waypoints. The final stage uses the three-finger manipulator to arrange several small dices in a specific pattern.

\begin{figure}[h]
\begin{center}
     \includegraphics[width=0.7\columnwidth]{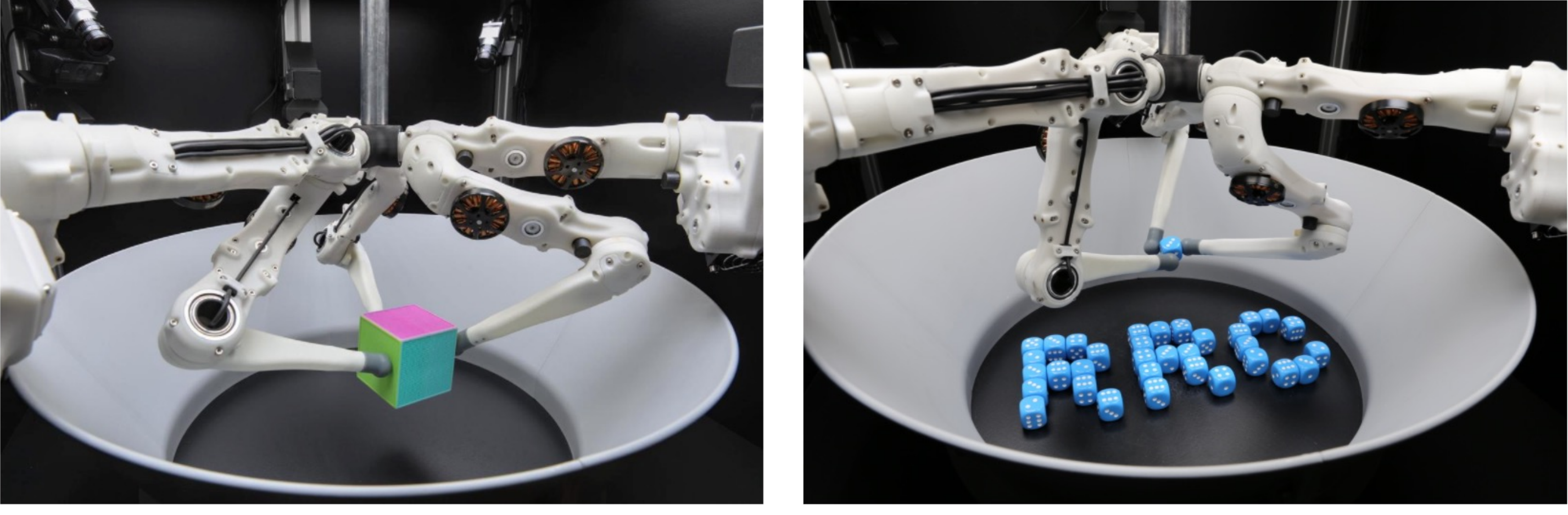}  
\end{center}
\caption{Environments from the RRC 2021. Left: Manipulating a cube along a sequence of waypoint poses, Right: Arranging the dices in a specific pattern.}
\label{fig:environments}
\end{figure}

\begin{figure}[t!]
\begin{center}
     \includegraphics[width=1.0\columnwidth]{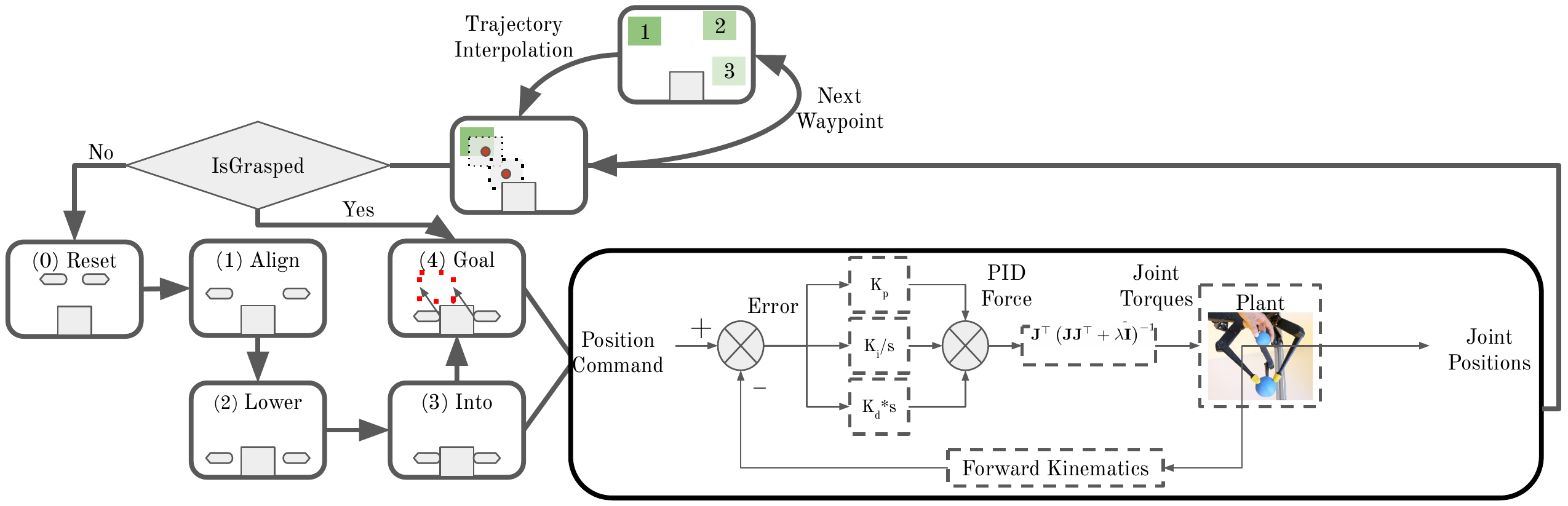}  
\end{center}
\caption{Representative state machine with trajectory interpolation}
\label{fig:method_rrc2021}
\end{figure}

In this report, we present a practical method for solving the first task of the challenge: manipulating the cube along a waypoint trajectory. 
We identify an issue in our previous approach \cite{funk2021benchmarking} which comprised of a Cartesian Position Controller with the Triangle Grasp: when the waypoint changes the controller accumulates error rapidly and the grasp used for manipulating the cube for the current waypoint might not be well suited for the next waypoint. To resolve this issue, we propose a fix using trajectory interpolation (see Figure \ref{fig:method_rrc2021}) -- we rediscretize the waypoints in a finer scale that leads to smoother motions with added grasp stability. We empirically test different grasping mechanisms and present an approach that achieved the runner-up position in Real Robot Challenge (RRC) 2021.

\section{Background}

\subsection{Problem setting}

We consider the first stage of the RRC 2021 challenge, which consists of a TriFinger manipulator tasked with moving the cube along a sequence of waypoints specified in the trajectory. The observations in the environment are composed of robot joint positions, velocities and torques along with the cube 6D pose and the desired goal position for the cube. The environment action interface allows us to control the robot via position control, torque control or both. The reward in the environment is the negative of distance to the current waypoint scaled by the range of motion available in each dimension. Precisely,
\begin{equation}
    r(o, g) = - \frac{||o_{xy,cube}-g_{x,y}||_2}{\text{range}_{xy}}+ \frac{|o_{z,cube}-g_{z}|}{\text{range}_{z}} 
\end{equation}
where $o$ is the observation, $g$ is the goal location of the cube, $||o_{xy,cube}-g_{x,y}||_2$ is the euclidean distance in the $x-y$ plane of the cube's current location and the goal location.

\subsection{Cartesian Position Control (CPC)}
CPC involves reducing the control problem in the joint space of all three fingers to performing control in the 3D $(x,y,z)$ Cartesian space of the end effector for each finger. We use Pybullet to obtain the Jacobian matrix $\mathbf{J} := \frac{\partial\mathbf{x}}{\partial\mathbf{q}}(\mathbf{q}, \dot{\mathbf{q}})$ and the damped pseudo inverse of $\mathbf{J}$ to ensure a good solution to joint velocity $\dot{\mathbf{q}} = \mathbf{J}^\top\left(\mathbf{J}\mathbf{J}^\top + \lambda\mathbf{I}\right)^{-1}\dot{\mathbf{x}}$. This is combined with the gravity compensation torques to command gravity-independent linear forces at each joint. These linear force commands are then used with a PID controller to construct position-based motion primitives. The controller takes a target Cartesian goal position and minimizes the primitive-specific error between the current and target position.

\subsection{Grasping Strategy}
We experiment with two different grasps: Three-Jaw Chuck Grasp and Triangle Grasp. Three-Jaw Chuck Grasp involves holding the cube using two opposite vertical faces by placing the end effector of one of the fingers on one side (as a thumb) and the other two end effectors on the opposite side (as the index and middle finger). Triangle Grasp entails placing the three end effectors on three of the four vertical faces of the cube such that they connect to form an equilateral triangle.

\subsection{Trajectory Interpolation}
The PID errors show steep deviations in values when directly switching from the current waypoint to the next waypoint. Inspired by traditional industrial robots, we implement linear trajectory interpolation in the Cartesian space, which samples $N$ equidistant points along the straight line connecting the current and next waypoint. This allows for smoother transitions between waypoints, thus improving the grasp stability due to reduced in-hand movement of the cube.

\section{Method}

\subsection{Pre-Stage: Solving the challenge in the simulated environment}
RRC 2021 challenge consists of a pre-stage challenge which is a simulated version of the Stage-1 challenge. This simulated environment allows for rapid prototying and testing and is naturally used as a testing ground for our method before we move on the real-world challenge. To solve the prestage challenge we rely on CPC for manipulation from the previous version of the challenge RRC 2020~\cite{funk2021benchmarking} and test various different grasping strategies like Triangle Grasp, Three-Jaw Chuck Grasp and Parallel grasp~\cite{cutkosky1990human} in this environment. This environment differs from the previous work~\cite{funk2021benchmarking} as previous work deals with manipulating the cube to a single goal locations whereas our task is manipulate the cube to a number of different goal locations. This necessitates grasping strategies that are robust to manipulator motions when reaching the immediate goal and consequent goal changes. Our experiments are built upon the open-source repository for the RRC 2020 challenge\footnote{https://github.com/cbschaff/benchmark-rrc}. 

Our experiments show that with the naive strategy of grasping the cube and manipulating with CPC, the robot frequently drops the cube when the active goal changes. This poor behaviour arises due to two major factors: (a) sudden jerk that the fingers experience due to a PID controller error accumulation and often-times attempts of fast-switching to active goal,  (b) a grasp that is poorly conditioned for different goals. These effects result in poor grasping and ultimately the cube slipping out.

We experiment with two strategies to solve this issue. First, we switched to Three-Jaw Chuck Grasp that allows for a more stable grasp across different goal locations when shifting from one goal position to another. Second, we implement a linear trajectory interpolation mechanism to generate more fine-grained intermediate waypoints or subgoals in the direction vector of the active goal, thus allowing for a smoother transition to the active goal using a PID controller. Although the Three-Jaw Chuck Grasp had high performance, we identified a potential shortcoming of Three-Jaw Chuck Grasp that it fails to regrasp dropped cube along the arena perimeter. This shortcoming of Three-Jaw Chuck Grasp can be rectified by using a Triangle Grasp, which empirically provides a more robust mechanism to lift the cubes near the outer perimeter of the arena.

\subsubsection*{Stage 1: Solving the challenge with Real Robot}
In this part of the challenge, we began experimentation on the real robot using our approach from pre-stage. After a few rounds of experiments and tuning cycles, we observed that, just as in simulation, the quality of the Three-Jaw Chuck Grasp degrades severely along the arena perimeter. Although we experimented with switching to a Triangle Grasp near the perimeter as proposed in pre-stage, Three-Jaw Chuck Grasp still negatively affected~\footnote{Example of failed Three-Jaw Chuck Grasp near the perimeter \href{https://youtu.be/RKZltgcjauY}{https://youtu.be/RKZltgcjauY}} our performance by failing to grasp dropped cube. This difference in performance across simulation and real robot might be attributed to a sim2real gap present in the domain. This motivated us to switch to the Triangle Grasp approach throughout the arena, which led to more stable grasps with rare drops. We then replaced Three-Jaw Chuck Grasp with Triangle Grasp and our the experiments. CPC-TG with Trajectory-Interpolation proved to be a surprisingly effective approach for this sequential goal reaching task.

\section{Results}
We evaluate a total of 8 real robot experiments with differing goal trajectories, spanning across three different robots.

\begin{singlespace}
\begin{center}
\begin{tabular}{ |c|c|c|c| } 
 \hline
 Test Environment & mean & median & stddev \\ 
 \hline
 Simulation (CPC-CG) & -7270.249 & -6113.044 & 2799.856 \\ 
 Real (CPC-CG) &-17071.5  & -17299.9 & -4106.18  \\ 
 \textbf{Real} (CPC-TG) & -9981.3 & -9718.5 & -1142.9\\ 
 \hline
\end{tabular}
\end{center}
\end{singlespace}

We observe Triangle Grasp to perform more robustly with fewer cube drops and obtaining higher cumulative reward. This is attributed to the sim2real gap where we observed that Triangle Grasp provided a more stable grasp qualitatively compared to the Three-Jaw Chuck Grasp (CPC-CG) that worked best in simulation. 
\begin{figure}[h]
\begin{center}
     \includegraphics[width=1.0\columnwidth]{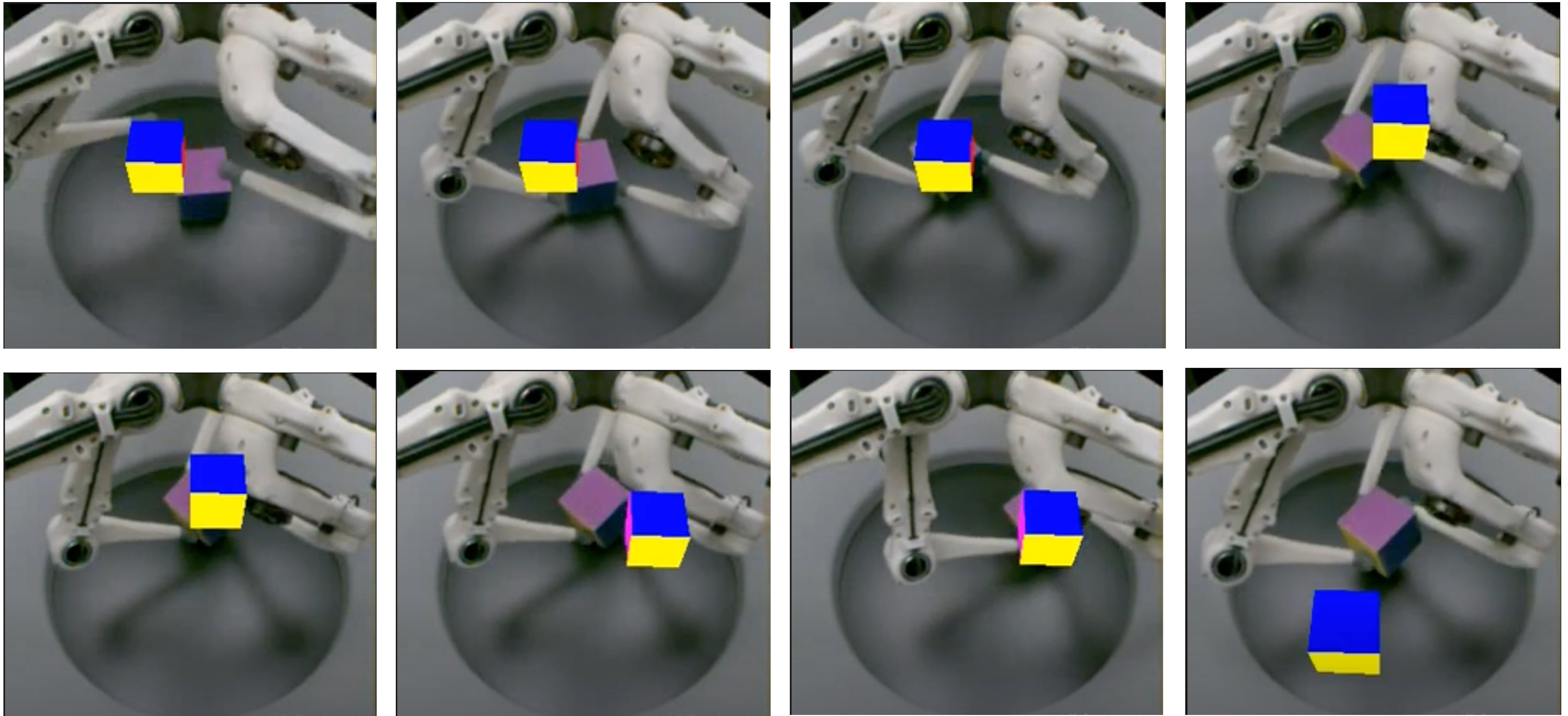}  
\end{center}
\caption{An example run of our method on an auto-generated sequence of goals. The objective is to match the goal position under any orientation. Video link: \href{https://youtu.be/dlOueoaRWrM}{https://youtu.be/dlOueoaRWrM}}
\label{fig:successful_run}
\end{figure}

\begin{figure}[h]
\begin{center}
     \includegraphics[width=1.0\columnwidth]{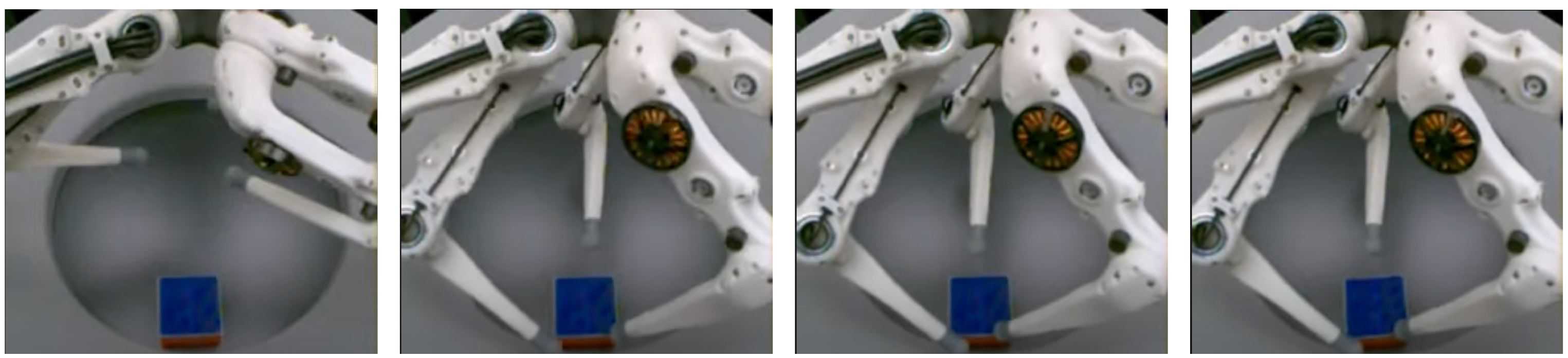}  
\end{center}
\caption{Grasping along the perimeter of the arena is difficult using the CPC along with the triangulated grasp. Video link:\href{https://youtu.be/RKZltgcjauY}{https://youtu.be/RKZltgcjauY} }
\label{fig:failed_pinch}
\end{figure}

Figure~\ref{fig:successful_run} shows our method's performance qualitatiely on an auto-generated sequence of tasks.  The method switches between goal smoothly and completes the task with a high cumulative reward compared to other methods submitted as part of the competition\footnote{https://real-robot-challenge.com/leaderboard. Our method was submitted under the alias decimalswift}. Figure~\ref{fig:failed_pinch} shows the failure case of our method where the robot is unable to grasp the object when it is close to the perimeter of the arena. We hypothesize that extrinsic dexterity can play a useful role in regrasping the cube in such situations and achieve even higher return. We leave this avenue to future work.

\paragraph{Discussion}
For the given task-environment, to achieve high rewards, the optimal strategy is to reach the current goal in the shortest path as fast as possible. Such optimal strategy requires having a stable grasp that minimizes cube drops and a controller that allows for quick transitions to goal. Overall our approach shows that manipulating a cube to reach various configurations can be effectively achieved through a simple PID controller with few key considerations of grasp and trajectory. Our method also transfers effectively to real robots requiring less than two hours of finetuning. It is interesting to draw comparisons on this particular task to learning-based methods like RL used in this environment~\cite{mccarthy2021solving}, which requires designing a shaped reward function and then dealing with the sim2real gap, both of which still require significant resources and human effort.



\clearpage


\bibliography{example}  

\end{document}